\documentclass{ecai}
\usepackage{graphicx}
\usepackage{latexsym}
\usepackage{inconsolata}
\usepackage{subfigure}
\usepackage{epstopdf}
\usepackage{amsmath}
\usepackage{booktabs}
\usepackage{multirow}
\usepackage{enumitem}
\usepackage{amssymb}
\usepackage{pifont}
\usepackage{bbding}
\usepackage{color}
\usepackage[dvipsnames]{xcolor}
\usepackage{float}
\usepackage{hyperref}

\begin{document}

\begin{frontmatter}

\title{Aspect-oriented Opinion Alignment Network for Aspect-Based Sentiment Classification}

\author[A]{\fnms{Xueyi}~\snm{Liu}}
\author[A]{\fnms{Rui}~\snm{Hou}}
\author[A]{\fnms{Yanglei}~\snm{Gan}} 
\author[A]{\fnms{Da}~\snm{Luo}} 
\author[A]{\fnms{Changlin}~\snm{Li}} 
\author[B]{\fnms{Xiaojun}~\snm{Shi}} 
\author[A]{\fnms{Qiao}~\snm{Liu}\thanks{Corresponding Author. Email: qiliu@uestc.edu.}}

\address[A]{University of Electronic Science and Technology of China}
\address[B]{China Academy of Electronics and Information Technology}

\begin{abstract}Aspect-based sentiment classification is a crucial problem in fine-grained sentiment analysis, which aims to predict the sentiment polarity of the given aspect according to its context. Previous works have made remarkable progress in leveraging attention mechanism to extract opinion words for different aspects. However, a persistent challenge is the effective management of semantic mismatches, which stem from attention mechanisms that fall short in adequately aligning opinions words with their corresponding aspect in multi-aspect sentences. To address this issue, we propose a novel Aspect-oriented Opinion Alignment Network (AOAN) to capture the contextual association between opinion words and the corresponding aspect. Specifically, we first introduce a neighboring span enhanced module which highlights various compositions of neighboring words and given aspects. In addition, we design a multi-perspective attention mechanism that align relevant opinion information with respect to the given aspect. Extensive experiments on three benchmark datasets demonstrate that our model achieves state-of-the-art results. The source code is available at \url{https://github.com/AONE-NLP/ABSA-AOAN}.
\end{abstract}

\end{frontmatter}

\section{Introduction}

The main purpose of aspect-based sentiment classification (ABSC) is to judge the sentiment polarity (positive, negative, neutral) \cite{liu2012sentiment,pontiki2016semeval} of aspect words in sentences expressing opinions. ABSC is an entity-level oriented and fine-grained challenge for sentiment analysis. To illustrate, consider the following sample sentence taken from the SemEval 2014 restaurant dataset:

\textit{Food is very good, though I occasionally wondered about freshness of raw vegetables in side orders}. 

In this sentence, the aspect are "\textit{food}" and "\textit{raw vegetables}", and the expected sentiment polarities of these aspects are intended to be positive and negative. Identifying the sentiment polarity of aspect is crucial for applications \cite{yadav2020sentiment,do2019deep} such as product review analysis, where understanding customers' opinions on specific aspects of a product can provide valuable insights for businesses.

\begin{figure}[]
\centerline{\includegraphics[scale=0.9]{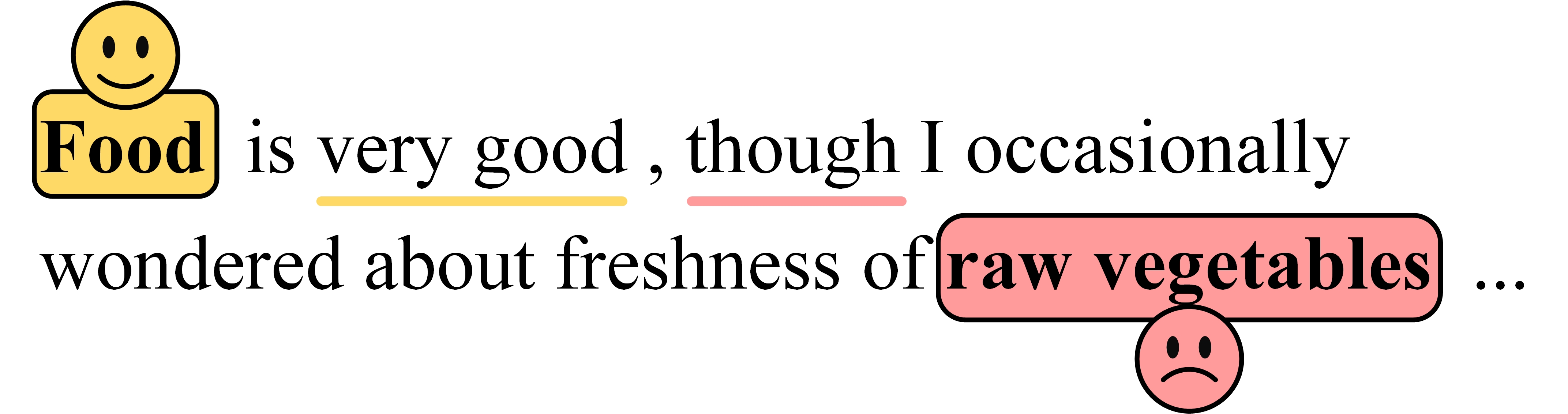}}
\caption{An example sentence contains two aspects but with opposite sentiment polarities from the restaurant reviews.} \label{example}
\end{figure}

To solve the aspect-based sentiment classification (ABSC) task, it is crucial to establish the semantic relationship between an aspect and its corresponding opinion words. Various recurrent neural networks (RNNs) \cite{nguyen2015phrasernn,tang-etal-2016-effective} have been proposed to learn representations directly from the left or/and right context with regard to the given aspect. However, prior studies have faced difficulties in accurately establishing semantic associations between aspect and long-distance context on sequential modeling. Therefore, attention mechanism have been widely adopted in ABSC tasks to model the correlations between aspect and context \cite{tang2016aspect,liu2018content,fan2018multi}. Unlike RNN-based models, attention mechanisms possess global modeling capability, which allows them to capture long-distance dependencies between aspect and context. However, attention mechanisms may not be effective when dealing with sentences containing multiple aspects. For instance, the given aspect "\textit{raw vegetables}" may be associated with the opinions words "\textit{very good}" and "\textit{though}" simultaneously. In such cases, attention mechanisms may struggle to align opinion words with their corresponding aspect, resulting in semantic mismatch \cite{liu2018content}.

To address this issue, various works introduce position information (e.g. fixed size window) \cite{gu2018position,zhou2020position} and proximity strategy (e.g. position weight decay) \cite{chen2017recurrent,li2018transformation}, which have proved that context words in closer proximity are more likely to be the actual opinion words of the aspect. However, these approaches may not encompass or emphasize all relevant opinion words, limiting the model's ability to fully comprehend the contextual meaning. As shown in Figure \ref{example}, the aspect "\textit{food}" has the opinion words "\textit{very good}" that are relatively close to the aspect and easy to capture its sentiment by the aforementioned methods. However, accurately calculating the sentiment of the aspect "\textit{raw vegetables}" requires the capture of a more comprehensive set of semantic information, including the opinion word "\textit{though}", which is much farther away from the aspect. Thus, the way prior works take position information into consideration may fall short in scenarios where the opinion words are distant from the aspect or convey complex semantic information. Therefore, the challenge remains on \textit{how to utilize attention mechanisms to accurately capture and match appropriate opinion words with respect to the given aspect}.

Drawing on the insights from recent studies on the critical role of different semantic compositionalities which can improve expressive ability and provide syntactic structure for natural language understanding \cite{socher2012semantic}, as well as the demonstrated effectiveness of span-level information in aspect-based sentiment classification (ABSC) \cite{zhou2019span,chen2022hierarchical}, we present a novel Aspect-oriented Opinion Alignment Network (AOAN) for aspect-based sentiment classification. Our proposed model addresses the issue of semantic mismatch by introducing a neighboring span enhanced module to highlights a variety of neighboring words compositions with respect to aspect. These compositions are used to emphasize different ranges of aspect neighboring words, providing flexibility to the contextual association between the aspect and neighboring words. To capture more comprehensive relevant opinion words based on different compositions of neighboring words, we then propose a multi-perspective attention module that utilizes abstract understanding representations to model multi-perspective sentiment representations of each aspect. This parallel attention mechanism improves the accuracy and comprehensiveness of capturing the relevant opinion words regarding the given aspect. Finally, the multi-perspective sentiment representations are combined by a global average pooling layer, which aggregates the information from all neighboring spans, providing a comprehensive representation of the overall sentiment expressed by the given aspect. The main contributions can be summarized as follows:

\begin{itemize}
    \item We propose an Aspect-oriented Opinion Alignment Network (AOAN), a novel framework designed to mitigate the semantic mismatch problem. This method is capable of exploiting contextual association of different neighboring spans and guarantees proper alignment between opinion words and the given aspect.
    \item Our propose AOAN employs a neighboring span enhanced module that highlights various compositions of neighboring words and given aspects, enabling the capture of more evident information related to a given aspect. Additionally, a multi-perspective attention module is designed to align comprehensive opinion information with the given aspect in a parallel way.
    \item We conduct extensive experiments on three benchmark datasets to evaluate the effectiveness of our approach. Experimental results demonstrate that our model outperforms the state-of-the-art methods, which confirms the efficacy of our proposed approach.
\end{itemize}

\section{Related Work}
Aspect-based sentiment classification (ABSC) is a fine-grained sentiment analysis task that focuses on extracting sentiment polarity towards a specific aspect within a given context. Early ABSC methods \cite{jiang2011target,kiritchenko2014nrc} relied on handcrafted features and failed to capture the intrinsic semantic associations between the given aspect and context.

Recently, various neural network-based approaches, such as Convolutional Neural Networks (CNNs) \cite{li2018transformation, kim2014convolutional}, Recurrent Neural Networks (RNNs) \cite{tang-etal-2016-effective,wang2016attention}, and Memory Networks \cite{tang2016aspect}, have been proposed to model the semantic relation between the aspect and context in an implicit way. For instance, Tang et al. \cite{tang-etal-2016-effective} introduced two LSTM-based models, namely TD-LSTM and TC-LSTM, which segmented the sentence into three parts: the preceding context words, the aspect, and the following context words. However, RNN-based and its variants methods face challenges in capturing long-distance contextual sentiment features when the aspect is far away from the opinion words, due to the limitation of sequential modeling.

With this in mind, researchers have deployed attention mechanisms for the aspect-based sentiment classification (ABSC) task to capture long-distance semantic features through global modeling. Tang et al. \cite{tang2016aspect} proposed a Deep Memory Network (MemNet) that utilizes an attention mechanism to explicitly capture the relevance of each contextual word with respect to the aspect and infer the sentiment polarity. However, the inherent defects of attention mechanisms cannot differentiate the correlations of contextual words with respect to the given aspect, leading to a semantic mismatch problem. 

To tackle the aforementioned issue, some works improved attention mechanisms by modeling a global perspective of sentence-level information \cite{liu2018content} or interaction between aspect and context \cite{ma2017interactive,huang2018aspect}. Liu et al. \cite{liu2018content} proposed a content attention-based aspect-based sentiment classification model (Cabasc) that captures crucial information about given aspects from a global perspective. Ma et al. \cite{ma2017interactive} introduced an interactive attention network (IAN) that generates the representations of target and context interactively. Huang et al. \cite{huang2018aspect} proposed an attention over attention networks (AOA) that learns attentions from both aspect-to-text and text-to-aspect, suggesting that opinion words are highly correlated with the aspect. However, they neglected the fact that the position information is also crucial for identifying the sentiment of the aspect.

Taking this into consideration, some researchers have introduced various position information and proximity strategies to improve the effectiveness of aspect-based sentiment classification (ABSC) models. Gu et al. \cite{gu2018position} proposed a position-aware bidirectional attention network (PBAN) that gives more attention to neighboring words of the aspect than words with long distances, while Zhou et al. \cite{zhou2020position} proposed a position-aware hierarchical transfer (PAHT) model that utilizes position information from multiple levels. Chen et al. \cite{chen2017recurrent} adopted a proximity strategy that assumes a closer opinion word is more likely to be the actual modifier of the target and designed a recurrent attention network (RAM) to counter irrelevant information using weight decay mechanisms. However, these approaches may not encompass or emphasize all relevant opinion words, limiting the model’s ability to fully comprehend the contextual meaning.

Another trend of research has explored the use of graph neural networks (GNNs) for modeling syntactic structures of sentence based on dependency trees. For instance, Zhang et al. \cite{zhang2019aspect} introduced aspect-specific graph convolutional networks (ASGCN) to handle aspect-level sentiment classification tasks. Tian et al. \cite{tian2021aspect} proposed a type-aware graph convolutional network (T-GCN) that utilizes an attentive layer ensemble to learn contextual information from different GCN layers. Li et al. \cite{li2021dual} proposed a dual graph convolutional network (DualGCN) that simultaneously took syntax structures and semantic correlations into consideration. Although syntactic-based methods have achieved promising results, the imperfect parsing performance and randomness of input sentences inevitably introduce noise through the dependency tree.

Prior works \cite{gu2018position,chen2017recurrent,li2018transformation} which take position information into consideration may fall short in scenarios where the opinion words are distant from the aspect or convey complex semantic information. In this paper, we do not rely on syntactic information and focus on the different compositions of aspect neighboring words, which provide comprehensive insights into the sentiment expressed.

\begin{figure*}[htp]
\centerline{\includegraphics[scale=0.73]{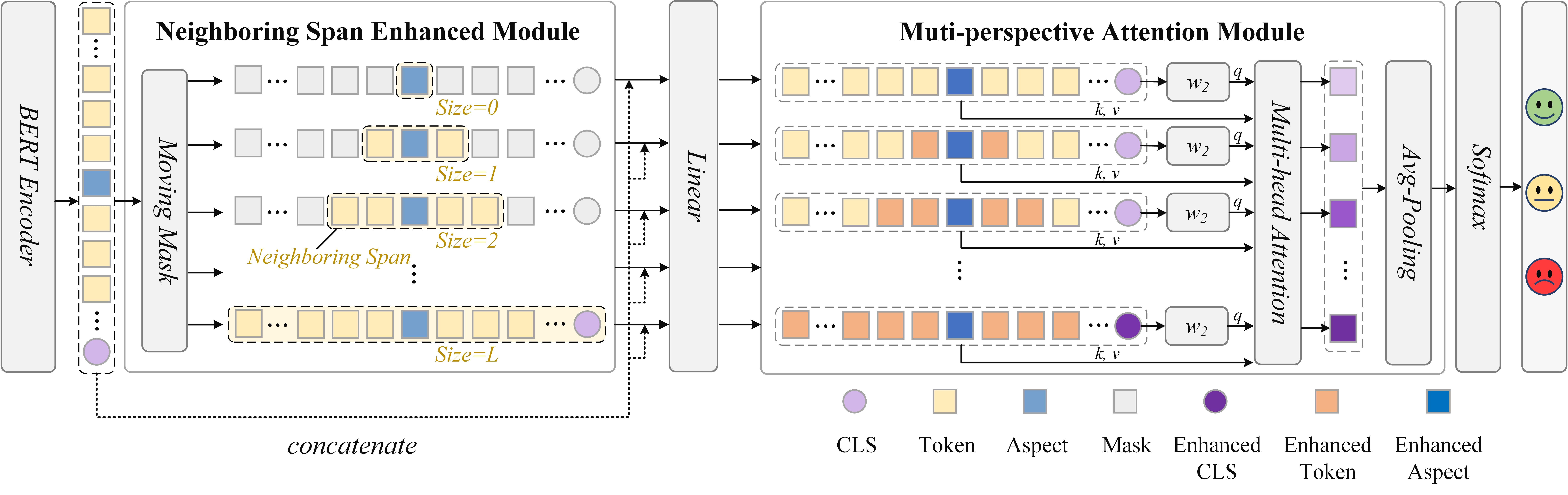}}
\caption{The overall architecture of AOAN, which is composed primarily of a neighboring span enhanced module highlights neighboring word spans of varying ranges, while a multi-perspective attention module captures relevant opinion words with a comprehensive view.} \label{procstructfig}
\end{figure*}

\section{Methodology}

\subsection{Overview}

We describe our model AOAN is this section, which has two main modules shown in Figure \ref{procstructfig}: a neighboring span enhanced module and a multi-perspective attention module. The neighboring span enhanced module highlights different compositions of neighboring words through multiple neighboring spans. The multi-perspective attention module captures relevant opinion words regarding the given aspect via multi-perspective sentiment representations. We will discuss each component in detail in the following sub-sections.

\subsection{Task Definition}\label{sub3.2}

The aim of our model is to predict the sentiment polarity of a given sentence towards a given aspect, based on the contextual information in the sentence. Specifically, let $S = \left\{w_1, \ldots, w_n\right\}$ represent a sentence comprising $n$ words, and let $A=\left\{w_{a+1}, \ldots, w_{a+m}\right\}$ denote the aspect mentioned in the sentence, consisting of $m$ words. Our goal is to accurately predict the sentiment polarity of the sentence $S$ towards the aspect $A$ from the set \{$positive$, $neutral$, $negative$\}.

\subsection{BERT Encoder}

To obtain contextual representations of sentences, we employ the pre-trained BERT language model \cite{devlin2019bert}, which maps sentences to input vector $H \in \mathbb{R}^{{(n+1)} \times d}$ including a class token and $n$ word tokens, where $d$ is the embedding dimension of the input vectors and $n$ is the length of input sentence. To capture aspect-aware context representations \cite{li2021dual,zhang2022ssegcn} of the sentence, we construct a sentence-aspect pair as "[CLS] \textit{Sentence} [SEP] \textit{Aspect} [SEP]" to serve as the input $S$ of BERT encoder. This process yields aspect-aware context representations $H= \{ h_{CLS},h_1,\dots,h_{n} \}$ of the sentence, which is the input of the subsequent modules and enable our model to effectively extract and comprehend the semantic meaning of the input.

\begin{equation}\label{eqn:1}
    H=BERT(S)
\end{equation}

\subsection{Neighboring Span Enhanced Module}

Taking into account the observation that words surrounding a given aspect tend to carry favourable information for sentiment polarity prediction \cite{gu2018position,zhou2020position, li2018transformation}, we construct multiple cohesive units, referred to as \textit{neighboring spans}, to model the tokens that precede and succeed the aspect, including the aspect itself. These neighboring spans encompass various ranges of tokens surrounding the aspect, each within a specific range of positions. Specifically, we employ a moving mask mechanism that first identifies the position of each token relative to the aspect. We then calculate the relative distance $d_i$ of each token with regard to the aspect, denoted as:

\begin{equation}\label{eqn:2}
    d_i=\begin{cases}P_{a+1}-P_i & i < a+1 \\ 0 & a+1 \leq i \leq a+m \\ P_i-(P_{a+1}+m) & i >a+m
    \end{cases}
\end{equation}

Here, $P_i$ denotes the position of $i$-th token, $i$ ranges from 0 to $n$, where $n$ is the length of the input sentence. $P_{a+1}$ denotes the start position of the aspect, and $m$ represents the length of the aspect. 

To model various composition of contextual neighboring words, we construct multiple neighboring spans based on the span size threshold $L$, which can range from 0 to $L$. This approach allows us to provide $L+1$ neighboring spans. The mask vector $V_i^l$ for each context word in a span of size $l \in [0,L]$ is defined as:

\begin{equation}\label{eqn:3}
V_i^l= \begin{cases}E & d_i \leq l \\ O & d_i>l\end{cases}
\end{equation}

where $E \in \mathbb{R}^{d}$ is the ones vector and $O \in \mathbb{R}^{d}$ is the zeros vector. $V_i^l$ denotes the mask vector of the $i$-th token for the span size $l$. We then obtain the $l$-th neighboring span representation $H_{span}^l \in \mathbb{R}^{{(n+1)} \times d}$, where the span size is $l$ as follows: 

\begin{equation}\label{eqn:4}
H_{span}^l=M^l \cdot H
\end{equation}

where the mask matrix $M^l=\{V_0^l, V_1^l, \ldots V_n^l\}\in \mathbb{R}^{{(n+1)} \times d}$ is composed of $n+1$ mask vectors $V_i^l$ and represents a span of size $l$ for the input sentence.

The neighboring span enhanced representation $H_{enhanced}^l \in \mathbb{R}^{{(n+1)} \times d}$ is then obtained by the concatenation of neighboring span representation and the original context representations $H$, and then mapped back to the unified $d$-dimensional semantic space via a shared linear transformation, denoted as:

\begin{equation}\label{eqn:6}
H_{enhance}^l=W_1(H_{span}^l \odot H)+b_1
\end{equation}

where $W_1 \in \mathbb{R}^{d \times 2d}$ and $b_1 \in \mathbb{R}^{d}$ are the weight and bias matrix in the linear transformation layer, and $\odot$ denotes concatenate operation. 

\subsection{Multi-perspective Attention Module}

In order to precisely capture opinion words and mitigate the problem of semantic mismatch, a multi-perspective attention module is utilized in this study. This module allows for parallel capturing sentiment information in neighboring span enhanced representations $H_{enhance}^l$ from various perspectives. Unlike other state-of-the-art approaches \cite{tang2016aspect,ma2017interactive,wang2021novel}, we utilize the abstract understanding representations $C^l \in \mathbb{R}^{d}$ transformed from $h^{l}_{CLS}$, as the query vector instead of the given aspect as the class token learns an abstract understanding of all the tokens in a sentence, which gives a comprehensive view. This ensures a more concentrated attention distribution in neighboring span, as it is believed that neighboring words may contain more valuable information for sentiment classification:

\begin{equation}\label{eqn:7}
C^l=f(W_2 \cdot h_{CLS}^l+b_2)
\end{equation}

\begin{equation}\label{eqn:8}
Q^l=W_Q \cdot C^l+b_Q
\end{equation}

\begin{equation}\label{eqn:9}
K^l=W_K \cdot H_{enhance}^l+b_K
\end{equation}

\begin{equation}\label{eqn:10}
V^l=W_V \cdot H_{enhance}^l+b_V
\end{equation}

where $W_2 \in \mathbb{R}^{d\times d}$ and $b_2 \in \mathbb{R}^{d}$ are the weight and bias matrix in a linear transformation layer, respectively. The activation function $Tanh$ is denoted as $f(\cdot)$. $W_Q \in \mathbb{R}^{d \times d}$, $W_K \in \mathbb{R}^{d \times d}$ and $W_V \in \mathbb{R}^{d\times d}$ are the weight matrices, $b_Q\in \mathbb{R}^{d}$, $b_K\in \mathbb{R}^{d}$ and $b_V\in \mathbb{R}^{d}$ are the bias matrices in three linear transformation layers.

The multi-head scaled dot-product attention is calculated as:

\begin{equation}\label{eqn:11}
y_{a}^l=W_h(head_1^l \odot head_2^l \odot \cdots \odot head_h^l)
\end{equation}

\begin{equation}\label{eqn:12}
head_{i}^l={Attention}(Q_i^l, K_i^l, V_i^l)
\end{equation}

\begin{equation}\label{eqn:13}
Attention(Q, K, V)=Softmax\left(\frac{Q K^T}{\sqrt{d_k}}\right) V
\end{equation}

where $\sqrt{d_k}$ is a scaling factor to avoid large values in the dot product between the query and key vectors, which can lead to instability in the softmax function. $Q_i^l$ , $K_i^l$ and $V_i^l$ denotes the query, key and value of the $i$-th head, respectively. The dimensions of $Q_i^l$ , $K_i^l$ and $V_i^l$ are $\frac{d}{h}$ and $h$ refers to the total number of heads. Furthermore, $W_h \in \mathbb{R}^{d \times d}$ represents a weight matrix to enable linear transformation.

Each attention output vector $y_{a}^l \in \mathbb{R}^{d}$ in  $[y_{a}^0,\cdots,y_{a}^L]$ are subjected to a linear transformation with shared parameters to produce three-dimensional sentiment representations $y_{s}^l \in \mathbb{R}^{3}$, defined as:

\begin{equation}\label{eqn:14}
y_{s}^l=W_3 \cdot y_{a}^l+b_3
\end{equation}

where $W_3 \in \mathbb{R}^{3 \times d}$ and $b_3 \in \mathbb{R}^{3}$ are the weight and bias matrix in the linear transformation layer, respectively.

To integrate the sentiment representations obtained from the diverse compositions of neighboring spans, we utilize a global average pooling layer, which effectively combines the multi-perspective sentiment representations $Y_s=\{y_s^0,...,y_s^L\} \in \mathbb{R}^{{(L+1)} \times 3}$, denoted as:

\begin{equation}\label{eqn:15}
y_{c}=AvgPool1d(Y_{s})
\end{equation}

where $y_c\in\mathbb{R}^{3}$ is the integrated representation and the $AvgPool1d(\cdot)$ is a \textit{1-d} average pooling layer with kernel size $L+1$.

\subsection{Sentiment Classifier}

The integrated representation $y_{c}$ can be utilized with a softmax function to predict sentiment polarity distribution:

\begin{equation}\label{eqn:16}
\hat{y}=Softmax\left(W_o y_{c}+b_o\right)
\end{equation}

where $\hat{y} \in \mathbb{R}^{3}$ represents the predicted sentiment polarity distribution, $W_o \in \mathbb{R}^{3 \times 3}$ and $b_o \in \mathbb{R}^{3}$ are learned parameters. To train the entire network, a loss function is minimized as follows:

\begin{table}
\centering
\caption{Statistics of the three experimental datasets.}
\label{tab:table1}
\begin{tabular}{@{}ccccccc@{}}
\toprule
\multirow{2}{*}{\textbf{Datasets}} & \multicolumn{2}{c}{\textbf{Positive}} & \multicolumn{2}{c}{\textbf{Negative}} & \multicolumn{2}{c}{\textbf{Neural}} \\ \cmidrule(l){2-7} 
                                   & \textbf{Train}     & \textbf{Test}    & \textbf{Train}     & \textbf{Test}    & \textbf{Train}    & \textbf{Test}   \\ \midrule
Laptop                             & 944                & 341              & 870                & 128              & 464               & 169             \\
Restaurant                         & 2164               & 728              & 807                & 196              & 637               & 196             \\
Twitter                            & 1561               & 173              & 1560               & 173              & 3127              & 346             \\ \bottomrule
\end{tabular}
\end{table}

\begin{equation}\label{eqn:17}
L(\hat{y}, y)=-\sum_{i=1}^N \sum_{j=1}^C y_i^j \log \left(\hat{y}_i^j\right)+\lambda\left(\sum_{\theta \in \Theta} \theta^2\right)
\end{equation}

where $y_i^j$ is the ground truth sentiment polarity, $C$ is the number of sentiment polarity categories, $\hat{y}_i^j$ represents the predicted sentiment probabilities, $\theta$ represents each parameter to be regularized, $\Theta$ is a collection of all parameters, $\lambda$ denotes the weight coefficient for $L_2$ regularization.

\section{Experiments}
\subsection{Datasets}
We perform experiments on three public standard benchmark datasets. The {\tt Restaurant} and {\tt Laptop} datasets from SemEval2014 task 4 \cite{pontiki2016semeval}, consist of reviews on the restaurant and laptop domains, respectively. The {\tt Twitter} dataset is constructed from Twitter posts \cite{dong2014adaptive}. All of these datasets include instances with three sentiment polarities: positive, neutral and negative. Moreover, each sentence in these datasets is annotated with corresponding aspects and their polarities. Statistics for the three datasets are presented in Table \ref{tab:table1}.

\subsection{Implementation Details}
For our experiments, we utilize 768-dimensional word embeddings initialized with \textit{bert-base-uncased English version} \footnote{https://huggingface.co/bert-base-cased} as the input. We employ the Adam optimizer with a learning rate of 2 × 10$^{-5}$. To address overfitting, we incorporate $L_2$ regularization with a regularization weight of $\lambda_2$ set to 1 × 10$^{-5}$. Hyper-parameters $L$ are swept over \{10, 10, 0\} on the {\tt Laptop}, {\tt Restaurant} and {\tt Twitter} datasets, respectively. The number of head $h$ in multi-head attention is set to 12. All the model weights are initialized from a uniform distribution. The model is trained in 20 epochs with a batch size of 16, and the maximum sequence length during training is set to 100. Furthermore, the model is trained on a single NVIDIA A100 Tensor Core GPU and an Intel Xeon Bronze 3104 CPU. We also conduct the T-test on multiple random seeds to ensure the robustness of our results.

\begin{table*}[]
\centering
\caption{Comparison of experimental results on three public datasets. Results that indicate a
statistically significant improvement with a p-value of less than 0.05 under the bootstrap paired t-test are marked with an "*".}
\label{tab:table2}
\begin{tabular}{@{}ccccccc@{}}
\toprule
\multirow{2}{*}{\textbf{Model}} & \multicolumn{2}{c}{\textbf{Laptop}}     & \multicolumn{2}{c}{\textbf{Restaurant}} & \multicolumn{2}{c}{\textbf{Twitter}}    \\ \cmidrule(l){2-7} 
                                & \textbf{Accuracy(\%)} & \textbf{F1(\%)} & \textbf{Accuracy(\%)} & \textbf{F1(\%)} & \textbf{Accuracy(\%)} & \textbf{F1(\%)} \\ \midrule
ATAE-LSTM                       & 68.70                 & -               & 77.20                 & -               & -                     & -               \\
MemNet                          & 72.37                 & -               & 80.95                 & -               & -                     & -               \\
RAM                             & 69.59                 & 64.61           & 78.30                 & 68.46           & 70.52                 & 67.95           \\
IAN                             & 72.10                 & -               & 78.60                 & -               & -                     & -               \\
Cabasc                          & 72.57                 & 67.06           & 79.37                 & 69.46           & 70.95                 & 69.00           \\
PBAN                            & 74.12                 & -               & 81.16                 & -               & -                     & -               \\
BERT-SPC                        & 78.99                 & 75.03           & 84.46                 & 76.98           & 73.55                 & 72.14           \\
AEN-BERT                        & 79.93                 & 76.31           & 83.12                 & 73.76           & 74.71                 & 73.13           \\
BERT4GCN                        & 77.49                 & 73.01           & 84.75                 & 77.11           & 74.73                 & 73.76           \\
T-GCN+BERT                      & 80.88                 & 77.03           & 86.16                 & 79.95           & 76.45                 & 75.25           \\
DualGCN+BERT                    & 81.80                 & 78.10           & 87.13                 & 81.16           & 77.40                 & 76.02           \\
AOIARN                    & 81.50                 & 78.20           & 83.40                 & 75.70           & 75.00                 & 73.80           \\
SSEGCN+BERT                     & 81.01                 & 77.69           & \textbf{87.31}        & 81.09           & 77.40                 & 76.02           \\ \midrule
\textbf{AOAN}                  & \textbf{82.45}*        & \textbf{79.09}*  & 86.43*                 & \textbf{81.19}*  & \textbf{77.46}*        & \textbf{76.54}*  \\ \bottomrule
\end{tabular}
\end{table*}

\subsection{Baseline Models}

To comprehensively evaluate our model, we compare it with state-of-the-art baselines. The models are briefly described as follows:
\begin{itemize}
    \item \textbf{ATAE-LSTM} \cite{wang2016attention}: Wang et al. combine the attention mechanism and LSTM to model the connection between aspect and context.
    \item \textbf{MemNet} \cite{tang2016aspect}: Tang et al. repeatedly use the attention mechanism to extract contextual features of aspect. The last attention output is used to predict sentiment polarity.
    \item \textbf{RAM} \cite{chen2017recurrent}: Chen et al. improve MemNet by representing memory with BiLSTM. At the same time, GRU is introduced to process the features of multi-layer attention mechanism.
    \item \textbf{IAN} \cite{ma2017interactive}: Ma et al. generate aspect and context representations through two LSTMs respectively, and learn aspect and context representations interactively.
    \item \textbf{Cabasc} \cite{liu2018content}: Liu et al. utilize two types of attention mechanisms to learn information of a given aspect from a global perspective and sequential correlations between the words and the aspect. 
    \item \textbf{PBAN} \cite{gu2018position}: Gu et al. utilize the relative positions between aspects and sentiment words to attend to informative parts of the input.
    \item \textbf{BERT-SPC} \cite{devlin2019bert}: The input of the pre-trained BERT model is "[CLS] + context + [SEP] + aspect + [SEP]". Finally, the sentiment polarity of aspect is predicted by the classifier "[CLS]".
    \item \textbf{AEN-BERT} \cite{song2019attentional}: Song et al. design an attention encoding network to extract features related to aspects in context, and improved performance by the pre-training BERT.
    \item \textbf{BERT4GCN} \cite{xiao2021bert4gcn}: Zhang et al. integrate the grammatical sequential features from the PLM of BERT and the syntactic knowledge from dependency graphs.
    \item \textbf{T-GCN} \cite{tian2021aspect}: Tian et al. utilize dependency types to distinguish different relations in the graph and uses attentive layer ensemble to learn the contextual information from different GCN layers.
    \item \textbf{DualGCN} \cite{li2021dual}: Li et al. design a SynGCN module and a SemGCN module with orthogonal and differential regularizers.
    \item \textbf{AOIARN} \cite{yang2022aspect}: Yang et al. adopts an interactive attention between aspect embeddings learned from GCN and opinion semantic embeddings learned from three bi-LSTMs.
    \item \textbf{SSEGCN} \cite{zhang2022ssegcn}: Zhang et al. model both semantic correlations and syntactic structures of a sentence for ABSC. 
\end{itemize}

\subsection{Main Results}
\begin{table}
\centering
\caption{Ablation Experiment Result on SemEval2014 datasets.}
\label{tab:my-table3}
\begin{tabular}{@{}rcccc@{}}
\toprule
\multicolumn{1}{c}{\multirow{2}{*}{\textbf{Model}}} & \multicolumn{2}{c}{\textbf{Laptop}}     & \multicolumn{2}{c}{\textbf{Restaurant}} \\ \cmidrule(l){2-5} 
\multicolumn{1}{c}{}                                & \textbf{Accuracy(\%)} & \textbf{F1(\%)} & \textbf{Accuracy(\%)} & \textbf{F1(\%)} \\ \midrule
\multicolumn{1}{l}{AOAN-NonAll}                      & 79.31                 & 75.94           & 85.45                 & 78.30           \\
\multicolumn{1}{l}{AOAN-NonSpan}                   & 79.00                 & 75.58           & 85.54                 & 78.47           \\
\multicolumn{1}{l}{AOAN-Aspect}                    & 80.41                 & 75.97           & 86.16                 & 78.91           \\
\multicolumn{1}{l}{AOAN-Maxpool}                   & 75.86                 & 71.15           & 81.25                 & 71.15           \\
\multicolumn{1}{l}{AOAN-Single(0)}                 & 80.09                 & 77.01           & 85.09                 & 77.70           \\
(1)                                                 & 80.25                 & 77.36           & 85.54                 & 79.14           \\
(2)                                                 & 79.15                 & 74.92           & 85.09                 & 77.75           \\
(3)                                                 & 79.62                 & 76.14           & 85.35                 & 77.69           \\
(4)                                                 & 79.94                 & 76.63           & 85.36                 & 77.24           \\
(5)                                                 & 79.62                 & 76.30           & 85.36                 & 78.34           \\
(6)                                                 & \textbf{81.19}        & \textbf{78.37}  & 85.09                 & 78.37           \\
(7)                                                 & 80.41                 & 77.75           & 85.36                 & 77.48           \\
(8)                                                 & 80.09                 & 76.49           & 85.18                 & 77.62           \\
(9)                                                 & 81.03                 & 78.73           & \textbf{86.34}        & \textbf{80.84}  \\
(10)                                                & 80.25                 & 77.66           & 84.82                 & 77.31           \\ \midrule
\multicolumn{1}{l}{\textbf{AOAN}}                  & \textbf{82.45}        & \textbf{79.09}  & \textbf{86.43}        & \textbf{81.19}  \\ \bottomrule
\end{tabular}
\end{table}

The assessment of Aspect-Based Sentiment Classification (ABSC) models is typically done using accuracy and macro-averaged F1-score. In Table \ref{tab:table2}, we present a comparison of our model with existing approaches. Our proposed AOAN model achieves state-of-the-art performance on the {\tt Laptop}, {\tt Restaurant} and {\tt Twitter} datasets with F1-scores of 79.09\%, 81.19\%, and 76.54\%, respectively. Although our approach falls slightly short of satisfactory accuracy on the {\tt Restaurant} dataset, it still performs well. These results indicate that our AOAN model can effectively capture accurate and comprehensive multi-perspective sentiment representations based on different neighboring span compositions to solve semantic mismatch. 

\begin{figure*}[htp]
\centering
\subfigure[Aspect \textit{food} on AOAN-NonAll]{
\label{heatmap31}
\includegraphics[scale=0.5]{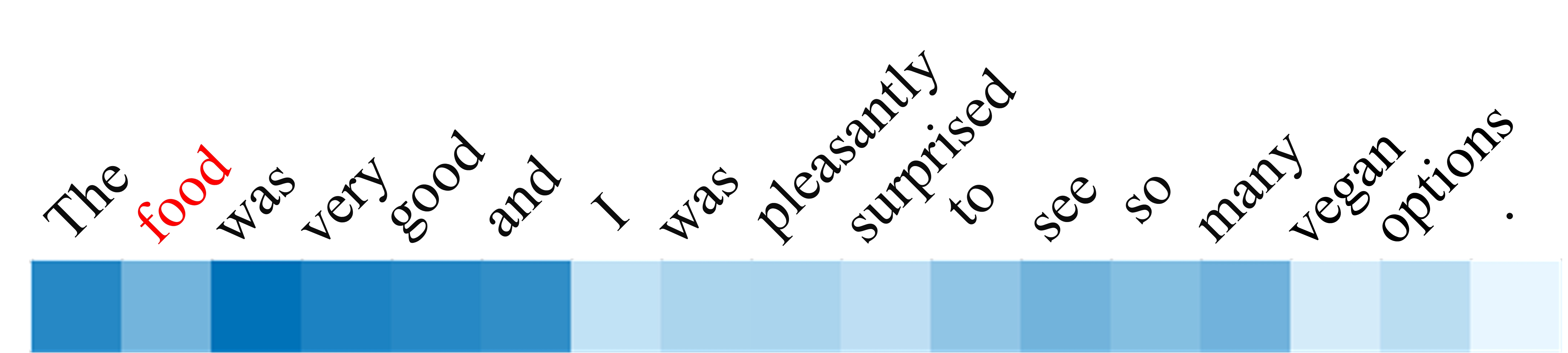}
}\subfigure[Aspect \textit{vegan options} on AOAN-NonAll]{
\label{heatmap41}
\includegraphics[scale=0.5]{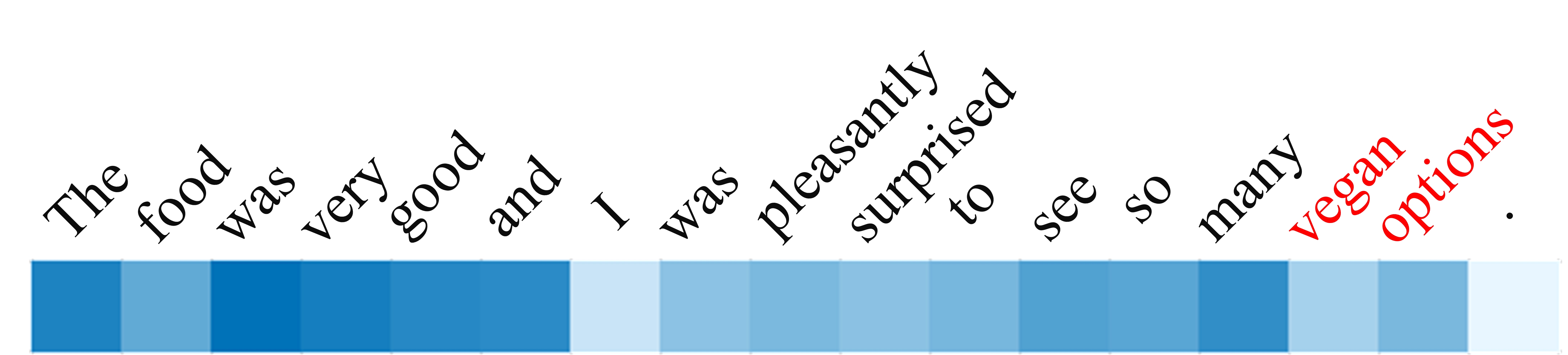}
}
\subfigure[Aspect \textit{food} on AOAN]{
\label{heatmap11}
\includegraphics[scale=0.5]{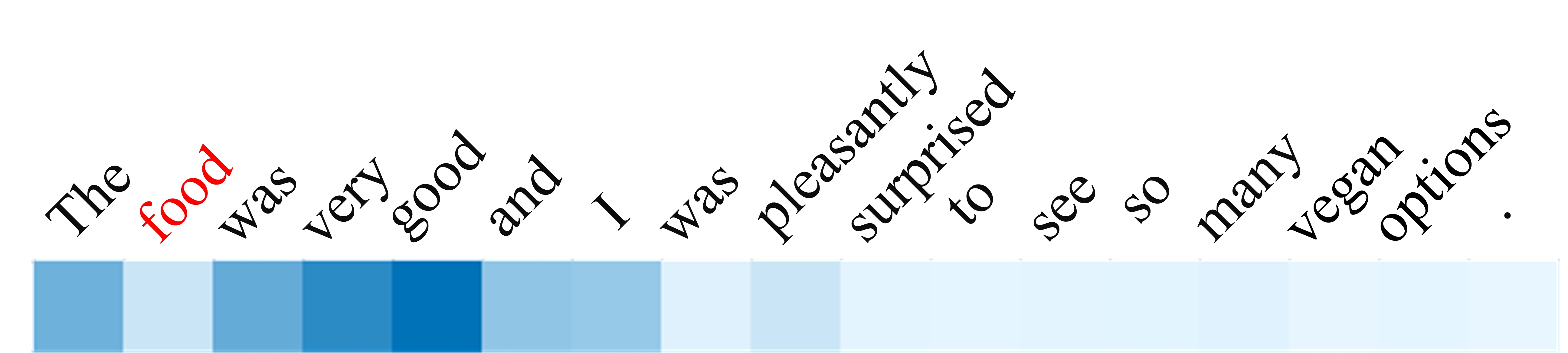}
}\subfigure[Aspect \textit{vegan options} on AOAN]{
\label{heatmap21}
\includegraphics[scale=0.5]{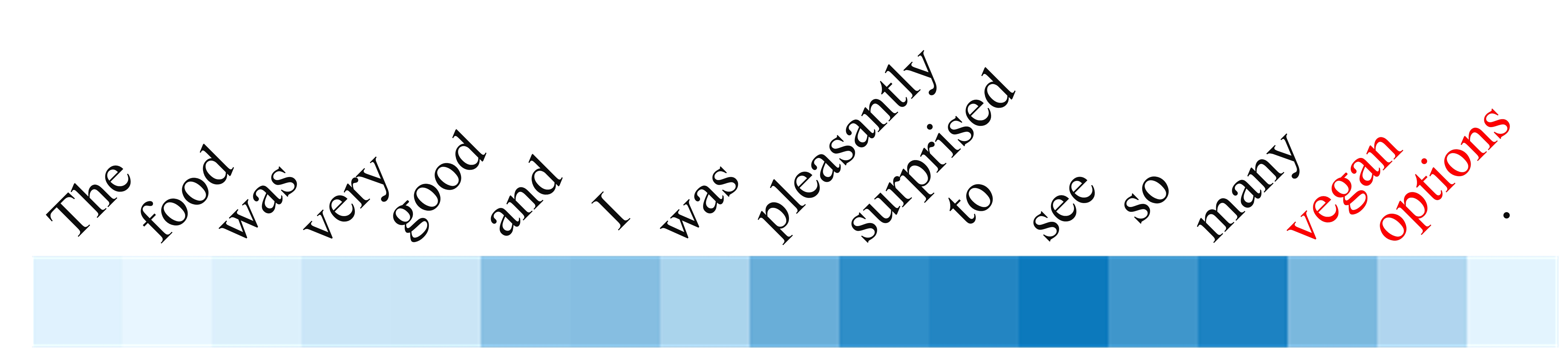}
}
\caption{The attention scores of AOAN-NonAll and AOAN model on {\tt Restaurant} dataset example.} 
\label{heatmap12}
\end{figure*}

Compared to semantic-based models, such as RAM, PBAN ,AEN-BERT and AOIARN, our AOAN model outperforms them with F1-score improvements of 0.89\%, 4.21\%, and 2.74\% on three datasets, respectively. Compared to RAM and PBAN which based on the position information and proximity strategy, one possible reason why our model performs better is that we use the BERT as the encoder which is effectively to extract contextual semantics. Another possible reason is that our model combines different neighboring word compositions and utilizes abstract understanding representations to capture important information from a global view while ignoring unimportant information makes it effective in aligning opinion words with their corresponding aspect. Furthermore, our model shows improved performance of 0.99\%, 0.03\%, and 0.52\% in F1-score on three datasets when compared to syntactic-based models such as BERT4GCN, T-GCN, DualGCN, and SSEGCN, without the need to incorporate any syntactic knowledge. In summary, the evaluation of our model shows its effectiveness and superiority in ABSC task, outperforming existing approaches without the need for complex syntactic structures.

\subsection{Ablation Study}

To investigate the effectiveness of the different modules in our AOAN model, we conducted extensive ablation studies on the SemEval2014 datasets, evaluating the impact of each module individually. The results of our experiments are presented in Table \ref{tab:my-table3}. We evaluated four different variants of our model: AOAN-NonAll, AOAN-NonSpan, AOAN-Aspect, AOAN-Maxpool and AOAN-Single(L). The AOAN-Single(L) model tests the model's ability to align opinion words with their corresponding aspect. Descriptions of each variants are detailed below:

\textbf{AOAN-NonAll} refers to a model that uses the output embeddings $H$ from the BERT encoder and the average vector of aspect representations $\{h_{a+1},\cdots,h_{a+m}\}$ as the input to an attention mechanism, which produces the final classification. 

\textbf{AOAN-NonSpan} removes the neighboring span enhanced module and instead uses the output embeddings $H$ from the BERT encoder as the input to the multi-perspective attention module.

\textbf{AOAN-Aspect} replaces the abstract understanding representation $C^l$ of AOAN with the average vector of aspect representations $\{h_{a+1}^l,\cdots,h_{a+m}^l\}$ in multiple neighboring span enhanced representations as the query for the multi-head attention mechanism.

\textbf{AOAN-Maxpool} replaces the global average pooling layer of AOAN with a max pooling layer to synthesize multi-perspective sentiment representations $Y_s$.

\textbf{AOAN-Single(L)} only uses the result of a single fixed span with size $L$, rather than combining the results of $L+1$ neighboring spans.

The experimental results demonstrate the contribution of each component to the overall effectiveness of the AOAN model. Specifically, utilizing aspect words to capture opinion words in the sentence (AOAN-NonAll) leads to a significant decrease in performance due to the lack of modules which align the opinion words with the given aspects. On the other hand, the effect of AOAN-NonSpan proves that the neighboring span enhanced module is effective to highlight the importance of neighboring words, resulting in improved performance. The comparison between AOAN-Aspect and AOAN shows that utilizing abstract understanding representations is a crucial clue to capture accurate and comprehensive opinion words.

\begin{figure}[htp]
\centering
\subfigure[{\tt Laptop} example]{
\label{heatmap01}
\includegraphics[scale=0.45]{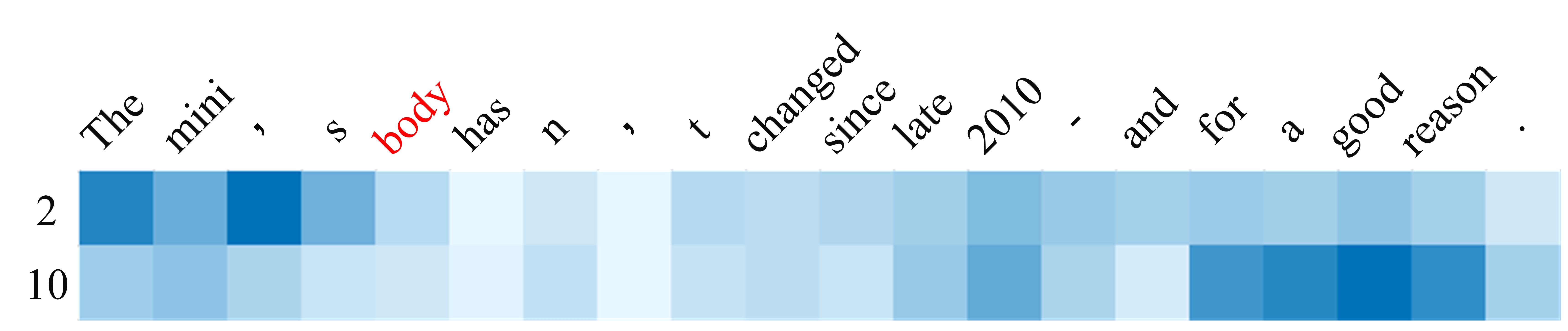}
}
\subfigure[{\tt Restaurant} example]{
\label{heatmap02}
\includegraphics[scale=0.5]{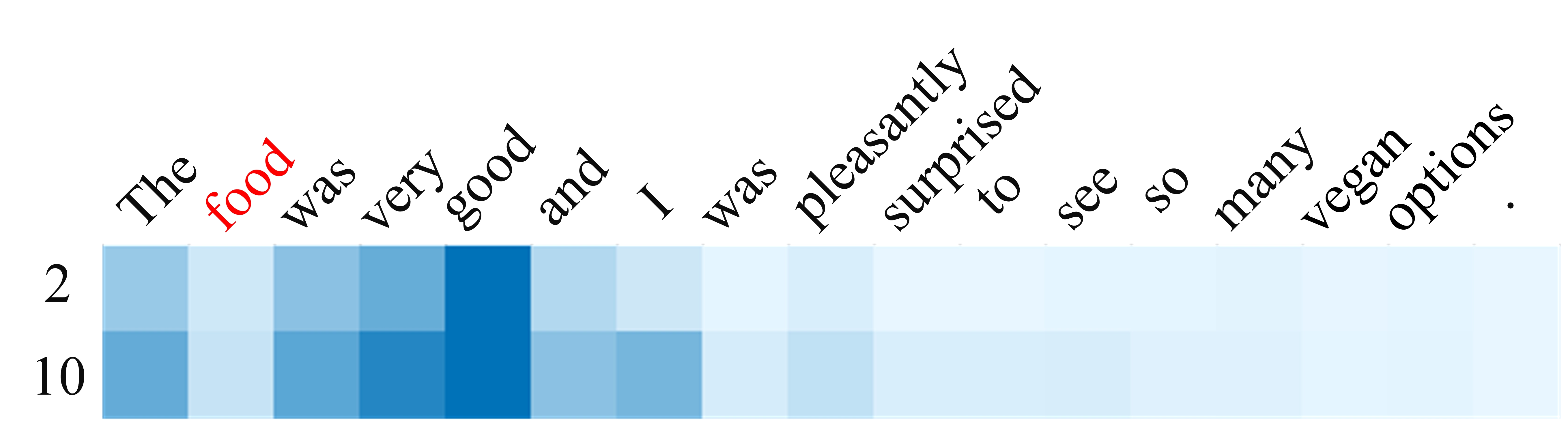}
}
\caption{The attention scores of AOAN model on {\tt Restaurant} and {\tt Laptop} datasets examples.} 
\label{heatmap0}
\end{figure}

Furthermore, the significant decrease in F1-score of 7.94\% and 10.04\% on {\tt Laptop} and {\tt Restaurant} datasets for AOAN-Maxpool suggests that each of the multi-perspective representations should be evenly utilized in the final classification. The lower performance of AOAN-Single(0) to AOAN-Single(10) in comparison to AOAN further supports the idea that combining different compositions can provide flexibility to the contextual association between the aspect and neighboring words. Overall, the results of the ablation experiments provide evidence for the effectiveness of each component of AOAN in improving the overall performance of the ABSC task.

\subsection{Visualization}

\begin{table*}[t]
\centering
\caption{Case studies of our AOAN model compared with MemNet, AEN-BERT and SSEGCN+BERT.}
\label{tab:my-table4}
\resizebox{\textwidth}{15mm}{
\begin{tabular}{@{}clcccc@{}}
\toprule
\textbf{\#} & \textbf{Sentence}                                                                  & \textbf{MemNet} & \textbf{AEN-BERT} & \textbf{SSEGCN} & \textbf{AOAN} \\ \midrule
1           & Not as fast as I would have expect for an \textcolor{red}{i5}.                                      & P{\scriptsize \XSolidBrush}             & N{\scriptsize \Checkmark}                & N{\scriptsize \Checkmark}                   & N{\scriptsize \Checkmark}             \\
2           & This newer netbook has no \textcolor{red}{hard drive} or \textcolor{red}{network lights}.                            & (N{\scriptsize \XSolidBrush},N{\scriptsize \XSolidBrush})           & (N{\scriptsize \XSolidBrush},N{\scriptsize \XSolidBrush})             & (N{\scriptsize \XSolidBrush},N{\scriptsize \XSolidBrush})                & (O{\scriptsize \Checkmark},O{\scriptsize \Checkmark})          \\
3           & Yes, they're a bit more expensive then typical, but then again, so is their \textcolor{red}{food}.  & O{\scriptsize \XSolidBrush}              & N{\scriptsize \XSolidBrush}                & N{\scriptsize \XSolidBrush}                   & P{\scriptsize \Checkmark}             \\
4           & Most of the \textcolor{red}{sandwiches} are made with \textcolor{red}{soy mayonaise} which is actually pretty good.  & (P{\scriptsize \Checkmark},P{\scriptsize \Checkmark})           & (O{\scriptsize \XSolidBrush},O{\scriptsize \XSolidBrush})             & (O{\scriptsize \XSolidBrush},P{\scriptsize\Checkmark})                 & (P{\scriptsize \Checkmark},P{\scriptsize \Checkmark})          \\
5           & But \textcolor{red}{dinner} here is never disappointing, even if the \textcolor{red}{prices} are a bit over the top. & (O{\scriptsize \XSolidBrush},N{\scriptsize \Checkmark})             & (P{\scriptsize \Checkmark},P{\scriptsize \XSolidBrush})               & (P{\scriptsize \Checkmark},N{\scriptsize \Checkmark})                  & (P{\scriptsize \Checkmark},N{\scriptsize \Checkmark})            \\ \bottomrule
\end{tabular}}
\end{table*}

To provide a clear understanding of the attention mechanism employed in our AOAN model, we present visualizations of the attention weights in two sample sentences in Figures \ref{heatmap12} and \ref{heatmap0}. The attention weights are computed as the average of all heads, and the intensity of the color represents the impact of the word on the sentence. To better illustrate, Figure \ref{heatmap12} shows the average of attention scores for each size of neighboring span, while Figure \ref{heatmap0} shows attention scores for different size of neighboring span in our AOAN model.

The sentence "\textit{The food was very good and I was pleasantly surprised to see so many vegan options.}" is taken from the {\tt restaurant} dataset with aspects "\textit{food}" and "\textit{vegan options}". Figure \ref{heatmap31} and \ref{heatmap41} show the attention scores of the AOAN-NonAll model, which uses aspect to capture opinion words. For the aspect "\textit{food}", unrelated words like "\textit{see}" and "\textit{many}" are given high weights, introducing noise. Similarly, for the "\textit{vegan options}" aspect, the irrelevant words "\textit{was very good}" receive the highest weights, leading to semantic mismatch. In contrast, our AOAN model (Figure \ref{heatmap11} and \ref{heatmap21}), with neighboring spans and multi-perspective attention, focuses attention around the aspect, aligning opinion words accurately.

In Figure \ref{heatmap0},  we show attention scores in span size 2 and 10 of the above {\tt Restautant} sentence and another example sentence, "\textit{The mini's body hasn't changed since late 2010 - and for a good reason.}" with the aspect "\textit{body}" taken from the {\tt laptop} dataset. In Figure \ref{heatmap01}, the span size of 2 fails to capture the long-distance context, while the relevant opinion words "\textit{for a good reason}" are aligned by the span size of 10. However, in Figure \ref{heatmap02}, the span size of 10 introduce more noises from irrelevant words such as "\textit{I}" and "\textit{pleasantly}" compared to span size of 2. The attention scores in different span size in our AOAN model reveal that the model can capture opinion words from different perspectives, which provide flexibility to the contextual association between the aspect and neighboring words. Therefore, the AOAN model can capture more comprehensive opinion words based on different compositions of neighboring words.

\subsection{Impact of Sentence Length}
To investigate the effectiveness of our proposed AOAN model in mitigating the problem of semantic mismatch, we considered the fact that longer sentences tend to contain more irrelevant words, thus exacerbating the issue. To this end, we defined a sentence whose length is greater than or equal to the average length as a long sentence and a sentence whose length is less than the average length as a short sentence. We then divided the test datasets of {\tt Laptop}, {\tt Restaurant}, and {\tt Twitter} into short and long sentences based on their respective average lengths \{21, 23, 26\}. We evaluated the accuracy of AEN-BERT which outperforms other semantic-based baselines, SSEGCN+BERT which outperforms other syntactic-based baselines, and our AOAN on short and long sentences.

\begin{figure}
\centering
\subfigure[long sentence]{
\label{longresult}
\includegraphics[scale=0.05]{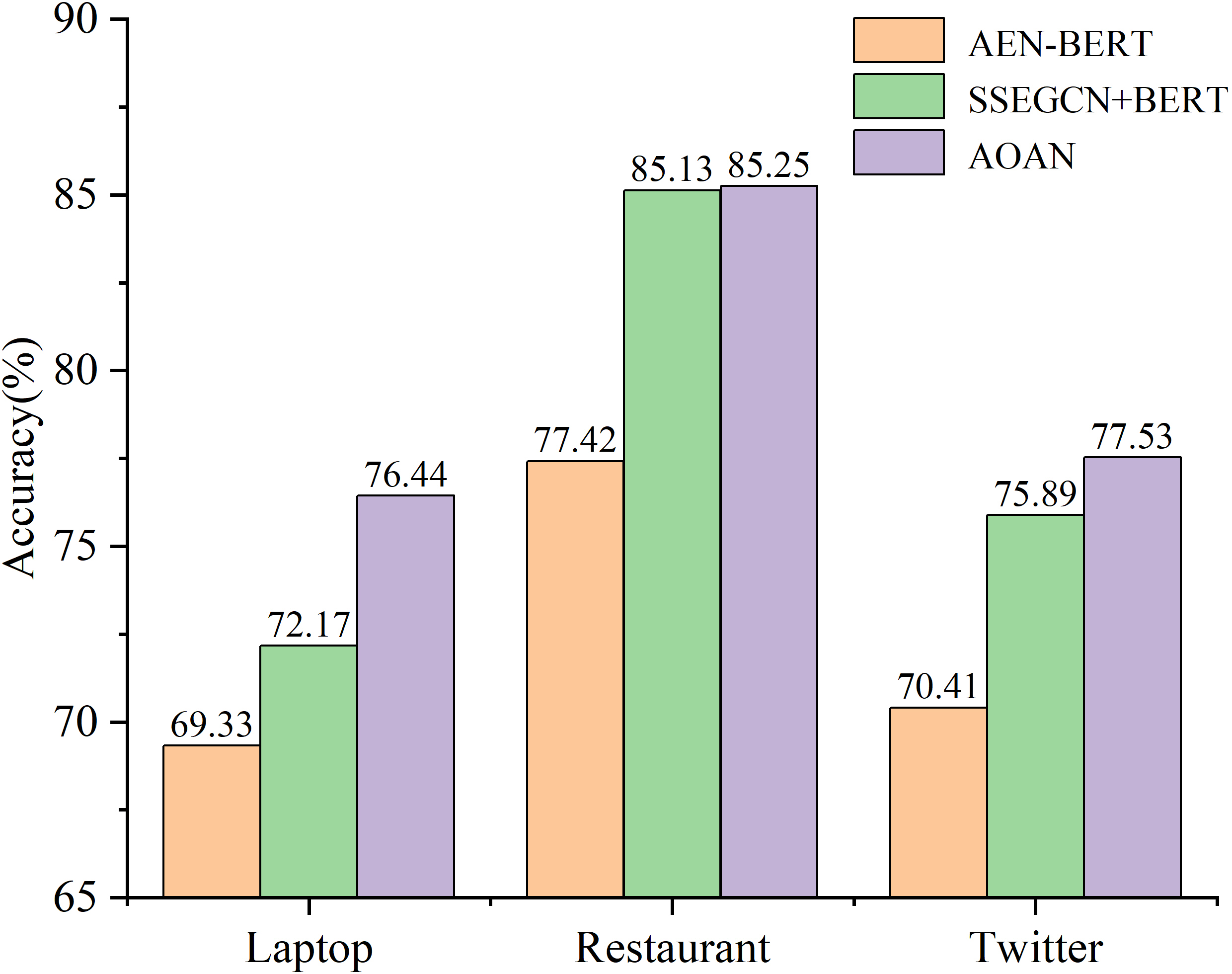}
}\subfigure[short sentence]{
\label{shortresult}
\includegraphics[scale=0.05]{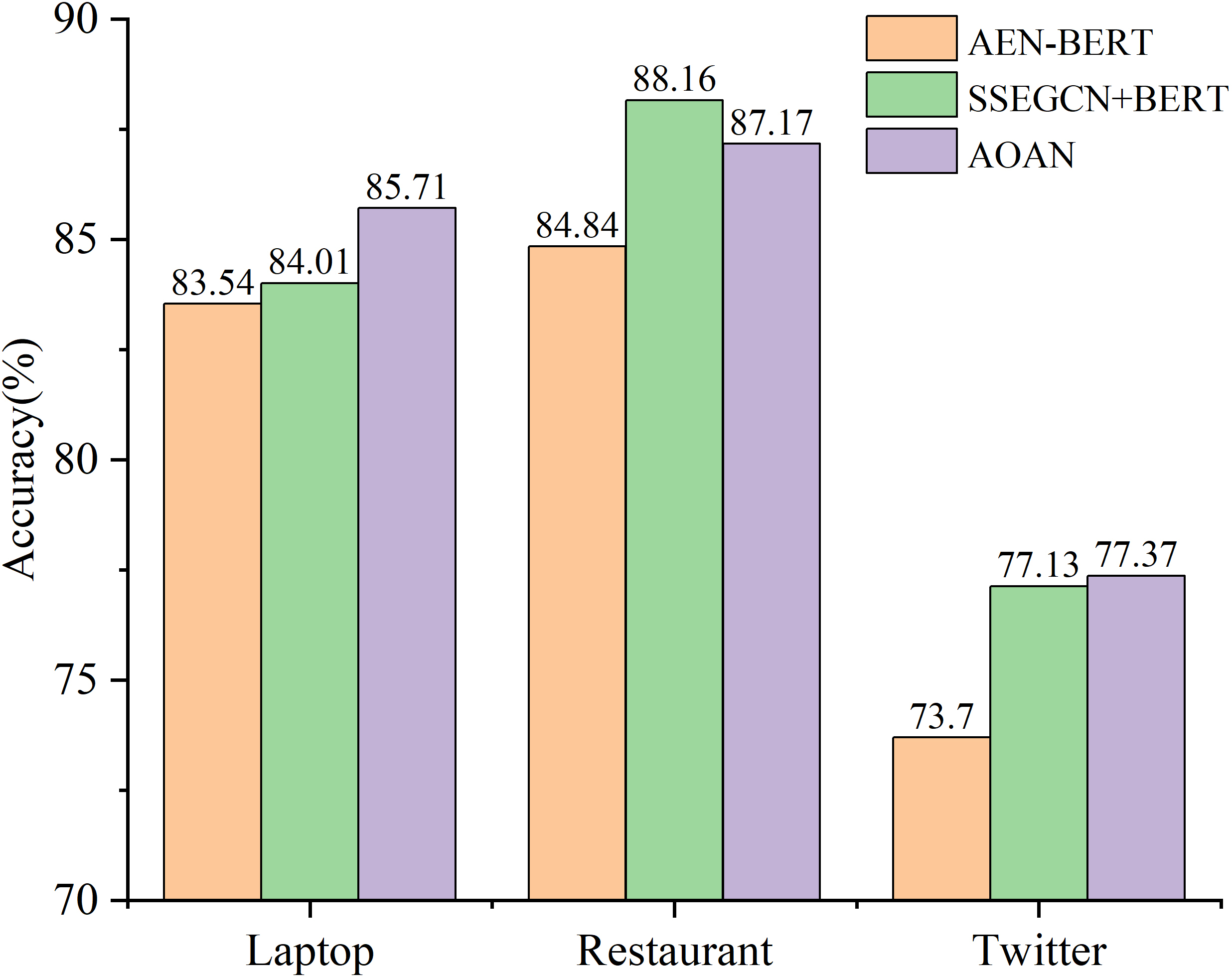}
}
\caption{Accuracies of our AOAN model on sentence of different length compared with AEN-BERT and SSEGCN+BERT.} 
\label{short and long result}
\end{figure}

As depicted in Figure \ref{longresult} and \ref{shortresult}, compared to AEN-BERT and SSEGCN-BERT, our model achieves better performance overall. Furthermore, our AOAN model achieves more significant improvement on long sentence compared with short sentence. Experimental results indicate that our proposed AOAN model is effective in mitigating the problem of semantic mismatch in long sentences and across varying sentence lengths, thereby improving the overall performance of sentiment analysis tasks.

\subsection{Case Study}

Table \ref{tab:my-table4} illustrates several case studies analyzed using different models, denoted by symbols P, N, and O for positive, negative, and neutral sentiment, respectively. Aspect words are highlighted in red for easy identification. In the first sentence, MemNet failed to consider the negative meaning of the word "\textit{not}" and made an incorrect prediction. In contrast, our AOAN model predicted neutral sentiment for the aspects "\textit{hard drive}" and "\textit{network lights}" in the second sentence, demonstrating its effectiveness in capturing overall sentiment. The third sentence, with no explicit sentiment expression, was correctly predicted by our AOAN model.

These examples highlight the importance of learning the integral semantics of a sentence for the ABSC task, and our AOAN successfully utilizes abstract understanding representations to capture comprehensive opinion information. In the last two sentences with different aspects, our AOAN accurately predicted the sentiment, showcasing its ability to align opinion words with their corresponding aspect and address the problem of semantic mismatch.

\section{Conclusions}
In this paper, we propose an AOAN to address the semantic mismatch problem in ABSC task. Specifically, we first design a neighboring span enhanced module, which highlights a variety of neighboring words compositions with respect to aspect for better capturing the accurate relevant opinion words. Then, we design a multi-perspective attention module to utilize abstract understanding to parallel model the multi-perspective sentiment representations that improves the accuracy and comprehensiveness of the capturing the relevant opinion words regarding the given aspect. Experimental results on three datasets achieve the state-of-the-art performance which demonstrate the effectiveness of our approach.

\ack We would like to thank the anonymous reviewers for their helpful discussion and feedback. This work was supported by the National Science Foundation of China (U19B2028, U22B2061) and the National Science and Technology Major Project of the Ministry of Science and Technology of China (2022YFB4300603).

\bibliography{ecai}
\end{document}